\documentclass[10pt,twocolumn,letterpaper]{article}

\usepackage{cvpr}              % To produce the CAMERA-READY version
\usepackage{graphicx}
\usepackage{amsmath}
\usepackage{amssymb}
\usepackage{booktabs}

\usepackage{bbm}
\usepackage{amsfonts}
\usepackage{color}
\usepackage{multirow}
\usepackage{listings}
\usepackage{newfloat}
\usepackage{algorithm}
\usepackage{algorithmic}
\usepackage{pifont}
\usepackage[pagebackref,breaklinks,colorlinks]{hyperref}

\usepackage[capitalize]{cleveref}
\crefname{section}{Sec.}{Secs.}
\Crefname{section}{Section}{Sections}
\Crefname{table}{Table}{Tables}
\crefname{table}{Tab.}{Tabs.}

\def\cvprPaperID{842} % *** Enter the CVPR Paper ID here
\def\confName{CVPR}
\def\confYear{2023}

\begin{document}

%%%%%%%%% TITLE - PLEASE UPDATE
\title{\emph{MagicNet}: Semi-Supervised Multi-Organ Segmentation via Magic-Cube Partition and Recovery}

\author{Duowen Chen\textsuperscript{1} \quad Yunhao Bai\textsuperscript{1} \quad Wei Shen\textsuperscript{2} \quad Qingli Li\textsuperscript{1} \quad Lequan Yu\textsuperscript{3} \quad Yan Wang\textsuperscript{1}\footnotemark[1]\\
\normalsize{\textsuperscript{1}Shanghai Key Laboratory of Multidimensional Information Processing, East China Normal University}\\
\normalsize{\textsuperscript{2}MoE Key Lab of Artificial Intelligence, AI Institute, Shanghai Jiao Tong University~~}
\normalsize{\textsuperscript{3}The University of Hong Kong}\\
{\tt\small duowen\underline{ }chen@hotmail.com, yhbai@stu.ecnu.edu.cn, wei.shen@sjtu.edu.cn,}\\
{\tt\small qlli@cs.ecnu.edu.cn, lqyu@hku.hk, ywang@cee.ecnu.edu.cn}\\
}
\maketitle
 \renewcommand*{\thefootnote}{\fnsymbol{footnote}}
 \setcounter{footnote}{1}
 \footnotetext{Corresponding Author.}
 \renewcommand*{\thefootnote}{\arabic{footnote}}
 \renewcommand*{\thefootnote}{\fnsymbol{footnote}}

%%%%%%%%% ABSTRACT
\begin{abstract}

We propose a novel teacher-student model for semi-supervised multi-organ segmentation. In teacher-student model, data augmentation is usually adopted on unlabeled data to regularize the consistent training between teacher and student. We start from a key perspective that \textbf{fixed relative locations} and \textbf{variable sizes} of different organs can provide distribution information where a multi-organ CT scan is drawn. Thus, we treat the prior anatomy as a strong tool to guide the data augmentation and reduce the mismatch between labeled and unlabeled images for semi-supervised learning. More specifically, we propose a data augmentation strategy based on partition-and-recovery N$^3$ cubes cross- and within- labeled and unlabeled images. Our strategy encourages unlabeled images to learn organ semantics in relative locations from the labeled images (cross-branch) and enhances the learning ability for small organs (within-branch). 
{For within-branch, we further propose to refine the quality of pseudo labels by blending the learned representations from small cubes to incorporate local attributes.}
Our method is termed as MagicNet, since it treats the CT volume as a magic-cube and N$^3$-cube partition-and-recovery process matches with the rule of playing a magic-cube. Extensive experiments on two public CT multi-organ datasets demonstrate the effectiveness of MagicNet, and noticeably outperforms state-of-the-art semi-supervised medical image segmentation approaches, with +7\% DSC improvement on MACT dataset with 10\% labeled images. Code is available at \url{https://github.com/DeepMed-Lab-ECNU/MagicNet}.

\end{abstract}

\begin{figure}[t]
\begin{center}
    \includegraphics[width=1\linewidth]{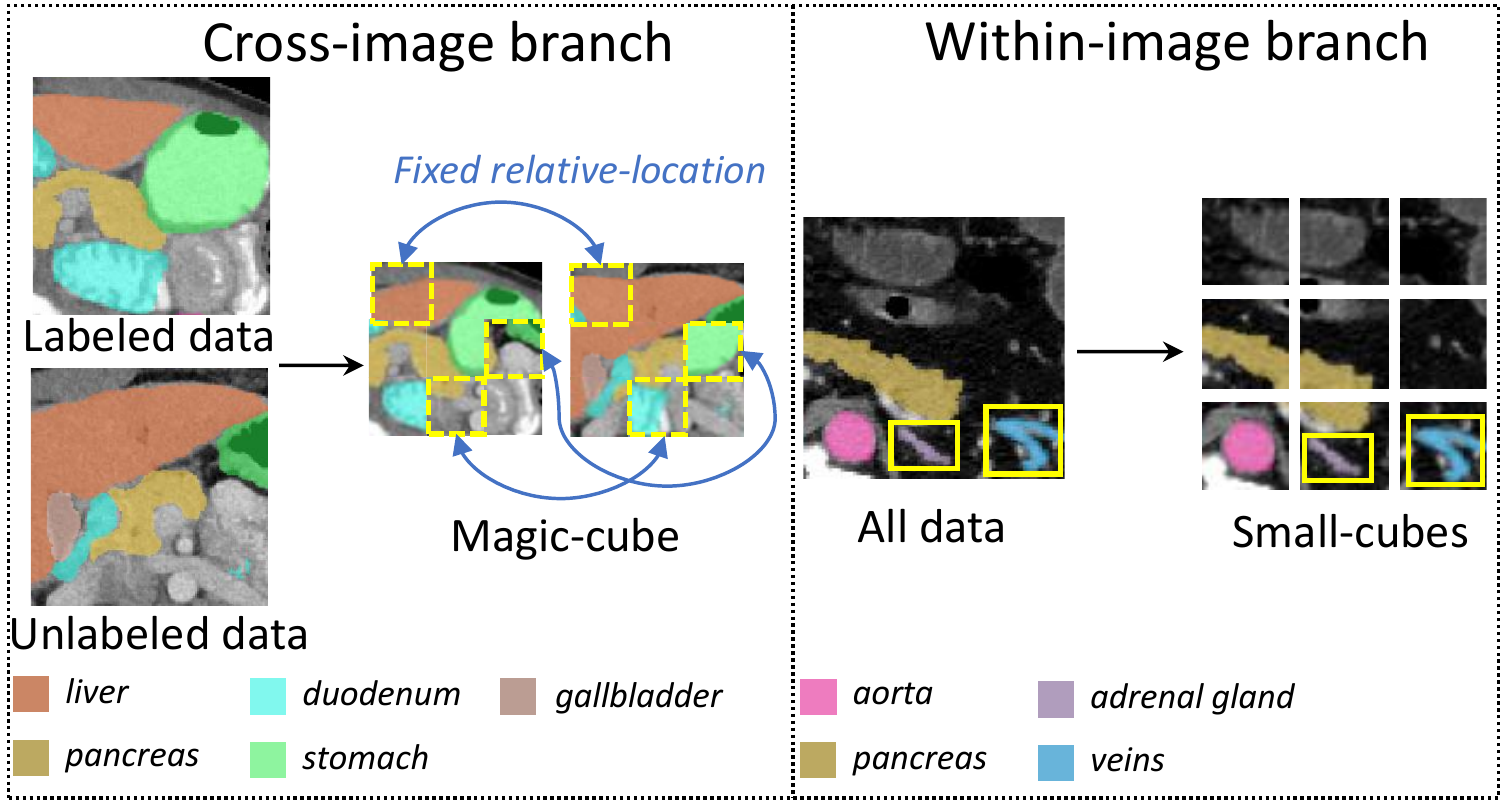}
\end{center}
\vspace{-1.3em}
\caption{ 
Two data augmentation strategies in MagicNet. \textbf{Left}: Although labeled and unlabeled images are not aligned, the latter can be regarded as a shifted version of the former. Co-shift of cubes transfers organ semantics in relative locations from the labeled data to unlabeled data. \textbf{Right}: Segmenting small organs from original images are difficult due to the cluttered background. Small cubes mitigate the impact from background and focus more on local attributes.} 
\label{fig:aug}
\vspace{-1.3em}
\end{figure}

%%%%%%%%% BODY TEXT
\section{Introduction}
\label{sec:intro}
Abdominal multi-organ segmentation in CT images is an essential task in many clinical applications such as computer-aided intervention \cite{Tang2019clinically,Wang2019abdominal}. But, training an accurate multi-organ segmentation model usually requires a large amount of labeled data, whose acquisition process is time-consuming and expensive. Semi-supervised learning (SSL) has shown great potential to handle the scarcity of data annotations, which attempts to transfer mass prior knowledge learned from the labeled to unlabeled images. SSL attracts more and more attention in the field of medical image analysis in recent years.

Popular SSL medical image segmentation methods mainly focus on segmenting a single target or targets in a local region, such as segmenting pancreas or left atrium \cite{Li2020shape,Luo2021DTC,You2022simcvd,Yu2019uncertainty,wu2022exploring,Bortsova2019semi,Wang2022semi,Fang202dmnet,CoraNet,Li2020transformation}. Multi-organ segmentation is more challenging than single organ segmentation, due to the complex anatomical structures of the organs, \emph{e.g.}, the fixed relative locations (duodenum is always located at the head of the pancreas), the appearances of different organs, and the large variations of the size. Transferring current SSL medical segmentation methods to multi-organ segmentation encounters severe problems. Multiple organs introduce much more variance compared with a single organ. Although labeled and unlabeled images are always drawn from the same distribution, due to the limited number of labeled images, it's hard to estimate the precise distribution from them \cite{Wang2019semi}. Thus, the estimated \textbf{distribution} between labeled and unlabeled images always suffers from \textbf{mismatch problems}, and is even largely increased by multiple organs. Aforementioned SSL medical segmentation methods lack the ability to handle such a large distribution gap, which requires sophisticated anatomical structure modeling. A few semi-supervised multi-organ segmentation methods have been proposed, DMPCT \cite{Zhou2019semi} designs a co-training strategy to mine consensus information from multiple views of a CT scan. UMCT \cite{Xia2020uncertainty} further proposes an uncertainty estimation of each view to improve the quality of the pseudo-label. Though these methods take the advantages of multi-view properties in a CT scan, they inevitably ignore the internal anatomical structures of multiple organs, resulting in suboptimal results.

Teacher-student model is a widely adopted framework for semi-supervised medical image segmentation \cite{NIPS2017_meanteacher}. Student network takes labeled images and unlabeled strongly augmented images as input, which attempts to minimize the distribution mismatch between labeled and unlabeled images from the model level. That is, data augmentation is adopted on unlabeled data, whose role is to regularize the consistent training between teacher and student. As mentioned, semi-supervised multi-organ segmentation suffers from large distribution alignment mismatch between labeled and unlabeled images. Reducing the mismatch mainly from the model level is insufficient to solve the problem. Thanks to the prior anatomical knowledge from CT scans, which provides the distribution information where a multi-organ CT scan is drawn, it is possible to largely alleviate the mismatch problem from the data level. 

To this end, we propose a novel teacher-student model, called \emph{MagicNet}, matching with the rule of playing a magic-cube. More specifically, we propose a partition-and-recovery $N^3$ cubes learning paradigm: (1) We partition each CT scan, termed as magic-cube, into $N^3$ small cubes. (2) Two data augmentation strategies are then designed, as shown in Fig.~\ref{fig:aug}. \emph{I.e.}, First, to encourage unlabeled data to learn organ semantics in relative locations from the labeled data, small cubes are mixed across labeled and unlabeled images while keeping their relative locations. Second, to enhance the learning ability for small organs, small cubes are shuffled and fed into the student network. (3) We recover the magic-cube to form the original 3D geometry to map with the ground-truth or the supervisory signal from teacher. Furthermore, the quality of pseudo labels predicted by teacher network is refined by blending with the learned representation of the small cubes. The cube-wise pseudo-label blending strategy incorporates local attributes \emph{e.g.}, texture, luster and boundary smoothness which mitigates the inferior performance of small organs.
 
The main contributions can be summarized as follows:
\begin{itemize}
    \item We propose a data augmentation strategy based on partition-and-recovery $N^3$ cubes cross- and within- labeled and unlabeled images which encourages unlabeled images to learn organ semantics in relative locations from the labeled images and enhances the learning ability for small organs.
    \item We propose to correct the original pseudo-label by cube-wise pseudo-label blending via incorporating crucial local attributes for identifying targets especially small organs.
    \item We verify the effectiveness of our method on BTCV \cite{landman2015miccai} and MACT \cite{densevnet2018TMI} datasets. The segmentation performance of our method exceeds all state-of-the-arts by a large margin, with 7.28\% (10\% labeled) and 6.94\% (30\% labeled) improvement on two datasets respectively (with V-Net as the backbone) in DSC.
\end{itemize}

\section{Related Work}
\subsection{Semi-supervised Medical Image Segmentation}
Semi-supervised medical image segmentation methods can be roughly grouped into three categories. (1) Contrastive learning based methods \cite{You2022simcvd, wu2022exploring}, which learn representations that maximize the similarity among positive pairs and minimize the similarity among negative pairs. (2) Consistency regularization based methods \cite{Bortsova2019semi,Luo2021DTC,Wang2022semi,Xia2020uncertainty,Yu2019uncertainty,Li2020shape,Fang202dmnet}, which attend to various levels of information for a single target via multi/dual-task learning or transformation consistent learning. (3) Self-ensembling/self-training based methods \cite{Yu2019uncertainty,CoraNet,Li2020transformation,Bai2017semi,Zhou2019semi}, which generate pseudo-labels for unlabeled images and propose several strategies to ensure the quality of pseudo-labels. But, most of these methods are mainly focusing on segmenting one target or targets in a local region or ROI, such as pancreas or left atrium \cite{You2022simcvd,CoraNet,Yu2019uncertainty,Luo2021DTC}, which encounters performance degradation when transferring to multi-organ segmentation, due to the lack of anatomical structure modeling ability.

\begin{figure*}[t]
\includegraphics[width=0.74\linewidth]{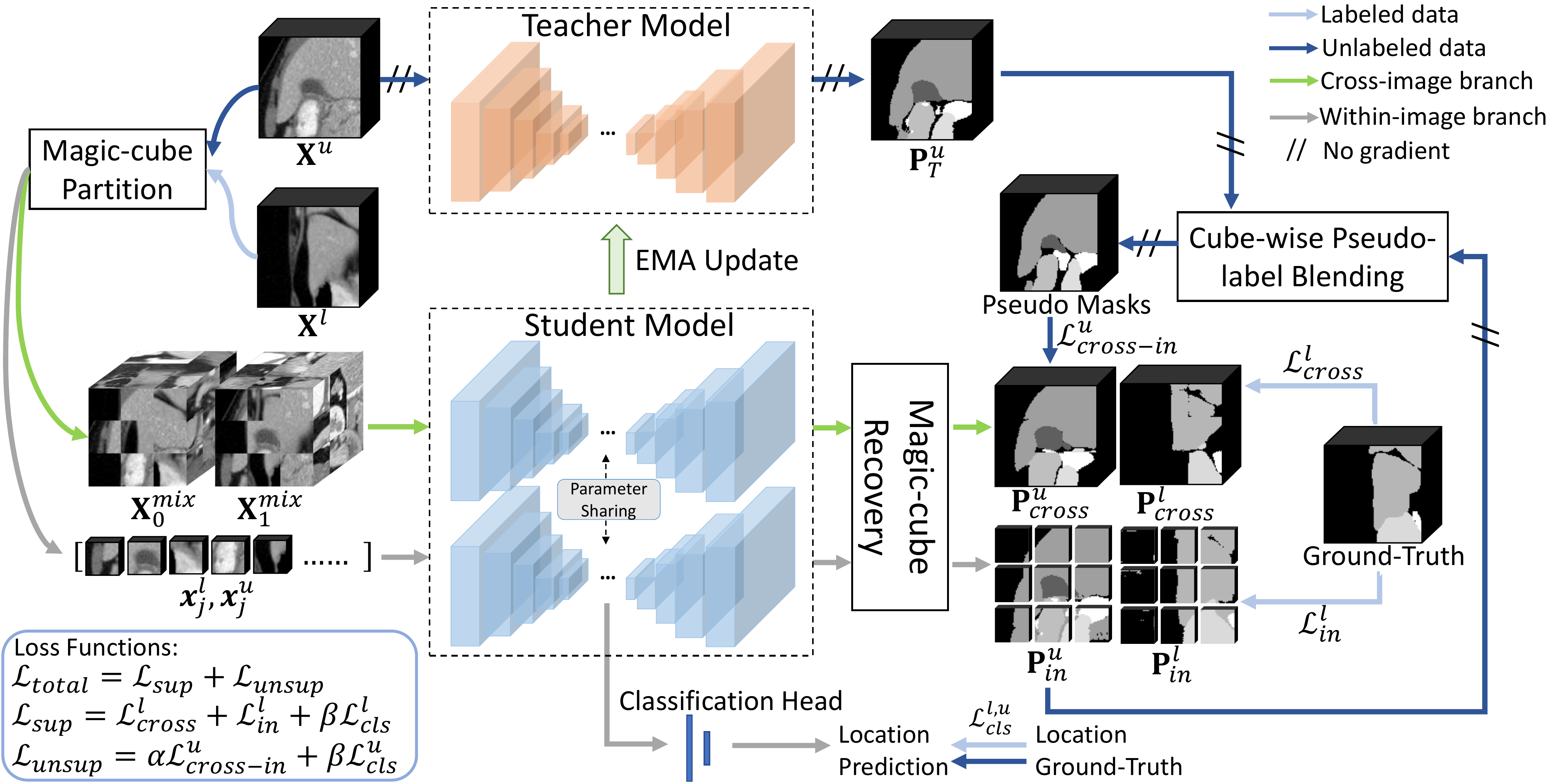}
\centering
\caption{The architecture of MagicNet. Two essential components are proposed: (1) magic-cube partition and recovery cross- and within-image (see Sec.~\ref{sec:magic-cube-PR} and Fig. \ref{fig:magic-cube partition and recovery}); (2) cube-wise pseudo-label blending, incorporating local attributes (see Sec.~\ref{sec:blending} and Fig. \ref{fig:bld}). 
}
\label{fig:framework}
\end{figure*}

\subsection{Semi-supervised Multi-organ Segmentation} 
Due to the large variations of appearance and size of different organs \cite{Wang2019abdominal,Holger2018an,Xie2020recurrent,Tang2021high}, multi-organ segmentation has been a popular yet challenging task. 
Only a few SSL methods specially for multi-organ segmentation have been proposed. DMPCT \cite{Zhou2019semi} adopted a 2D-based co-training framework to aggregate multi-planar features on a private dataset. UMCT \cite{Xia2020uncertainty} further enforces multi-view consistency on unlabeled data. These methods are pioneer works to handle semi-supervised multi-organ segmentation problem from the co-training perspective, which take advantages of the {multi-view property of a CT volume},
but the properties of multiple organs have not been well-explored. 

\subsection{Interpolation-based Semi-supervised Learning}
Interpolation-based regularization \cite{Yun_2019_ICCV,ijcai2019-504,zhang2018mixup} is quite successful in semi-supervised semantic segmentation. Methods such as Mixup \cite{zhang2018mixup} and CutMix \cite{Yun_2019_ICCV} are good at synthesizing new training samples and are widely used in semi-supervised learning. FixMatch \cite{sohn2020fixmatch} design a weak-strong pair consistency to simplify semi-supervised learning. MixMatch \cite{2019MixMatch} and ReMixMatch \cite{2019ReMixMatch} generated pseudo labels for unlabeled data and incorporate gradually unlabeled data with reliable pseudo-label into the labeled set. ICT \cite{ijcai2019-504} proposed a interpolated-based consistency training method based on Mixup. GuidedMix-Net \cite{2022GuidedMixNet} utilized Mixup \cite{zhang2018mixup} in semi-supervised semantic segmentation. Thus, we also compare our approach with some popular interpolated-based methods in Experiments.

\section{Method}

We define the 3D volume of a CT scan as $\textbf{X}\in\mathbb{R}^{W\times H\times L}$. The goal is to find the semantic label of each voxel $k\in \textbf{X}$, which composes into a predicted label map $\widehat{\textbf{Y}}\in\{0,1,...,C\}^{W\times H\times L}$, where $C=0$ indicates background class, and $C\neq 0$ represents the organ class. The training set $\mathcal{D}$ consists of two subsets: $\mathcal{D}=\mathcal{D}^l\cup \mathcal{D}^u$, where $\mathcal{D}^l = \{(\textbf{X}^l_i,\textbf{Y}^l_i)\}_{i=1}^N$ and $\mathcal{D}^u = \{\textbf{X}^u_i\}_{i=N+1}^{M+N}$. The training images $\textbf{X}^l$ in $\mathcal{D}^l$ are associated with per-voxel annotations $\textbf{Y}^l$ while those in $\mathcal{D}^u$ are not. In the rest of this paper, we denote the original and mixed CT scans as magic-cubes, and denote the partitioned small cubes as cubes for simplicity.

The overall framework of the proposed {MagicNet} is shown in Fig.~\ref{fig:framework}, consisting of a student network and a teacher network. The teacher network is updated via weighted combination of the student network parameters with exponential moving average (EMA)\cite{NIPS2017_meanteacher}. Designed in this architecture, our whole framework includes two branches: Cross-image branch and within-image branch. The details will be illustrated in the following sections. 

\subsection{Magic-cube Partition and Recovery}

\label{sec:magic-cube-PR}
Assume that a mini-batch $\mathcal{B}$ contains $n$ images $\textbf{X}^{\mathcal{B}}\in\mathbb{R}^{n\times W\times H\times L}$. For notational simplicity, let $n=2$, and $\mathcal{B}$ contains one labeled image $\textbf{X}^l$ and one unlabeled image $\textbf{X}^u$, which are randomly sampled from $\mathcal{D}^l$ and $\mathcal{D}^u$, respectively. We partition $\textbf{X}^l$ and $\textbf{X}^u$ into $N^3$ magic-cubes $\{\textbf{x}_{j}^u\}_{j=1}^{N^3}$ and $\{\textbf{x}_{j}^l\}_{j=1}^{N^3}$, where $\textbf{x}_{j}^l, \textbf{x}_{j}^u\in\mathbb{R}^{W/N\times H/N\times L/N}$, $j$ represents the relative location of the cube inside this magic-cube. This partition operation is called as $\mathcal{A}_{part}(\cdot)$.

\begin{figure*}[t]
\includegraphics[width=0.85\linewidth]{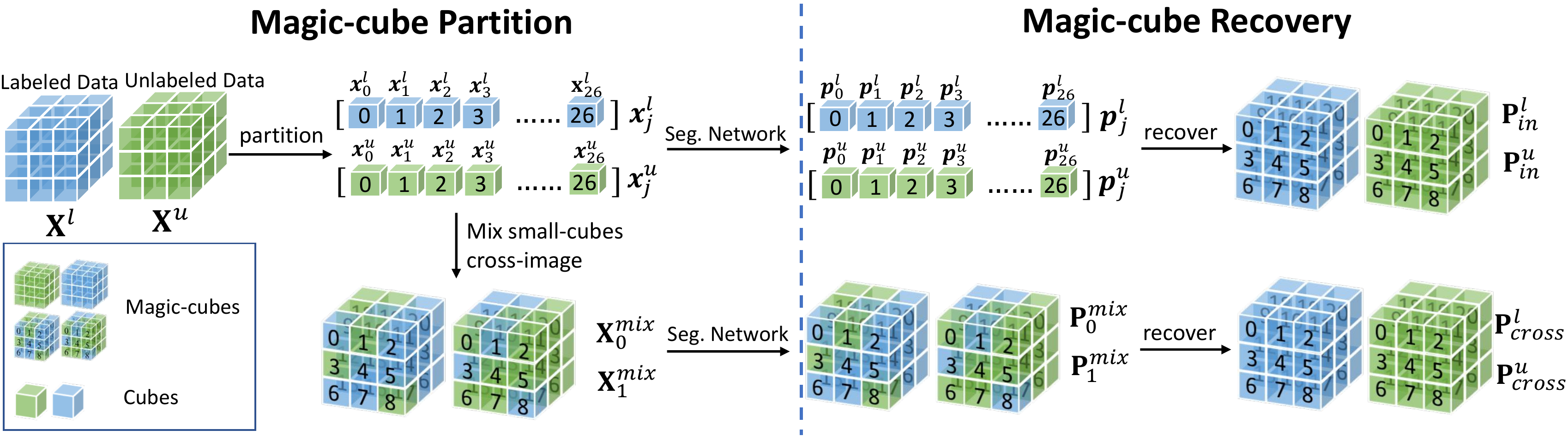}
\centering
\caption{Magic-cube partition and recovery. {Blue and green volumes represent labeled and unlabeled image, respectively. The number on cubes represents its relative-location in the original image.} }
\label{fig:magic-cube partition and recovery}
\vspace{-0.5em}
\end{figure*}

\noindent\textbf{Cross-image Partition and Recovery} 
To encourage labeled and unlabeled images to learn comprehensive common semantics from each other, we mix these cubes across all the labeled and unlabeled images in a mini-batch. As shown in Fig. \ref{fig:magic-cube partition and recovery}, these cubes $\textbf{x}_{j}^l$ and $\textbf{x}_{j}^u$ are mixed into two shuffled magic-cubes while keeping their original positions, which produce two interpolated images with mixed magic-cubes $\textbf{X}^{mix}_0, \textbf{X}^{mix}_1\in \mathbb{R} ^ {W\times H\times L}$. The magic-cube mixing cross-image operations are termed as $\mathcal{A}_{mix}^{cross}(\cdot)$. The mixed images are then fed into the student network $\mathcal{F}(\cdot;\mathbf{\Theta}^s)$, followed by a softmax layer $\sigma(\cdot)$, obtaining the prediction maps ${\textbf{P}}_0^{mix}, {\textbf{P}}_1^{mix}\in\mathbb{R} ^ {C\times W\times H\times L}$, where $C$ denotes the number of classes. Next, we recover ${\textbf{P}}^{l}_{cross}, {\textbf{P}}^{u}_{cross}\in \mathbb{R} ^ {C\times W\times H\times L}$ from ${\textbf{P}}^{mix}_0$ and ${\textbf{P}}^{mix}_1$ via recovering the magic-cubes back to the original positions in their original images, and this operation is denoted as $\mathcal{A}_{rec}^{cross}(\cdot)$. We can simply denote the whole process as\vspace{-0.7em}:

\begin{footnotesize}
\begin{align}
\label{eq:cross}
    {\textbf{P}}^{l}_{cross},{\textbf{P}}^{u}_{cross} = \mathcal{A}_{rec}^{cross}(\sigma(\mathcal{F}(\mathcal{A}_{mix}^{cross}(\mathcal{A}_{part}(\textbf{X}^\mathcal{B}));\mathbf{\Theta}^s))).
\end{align} 
\end{footnotesize}

\noindent Here, ${\textbf{P}}^{l}_{cross}, {\textbf{P}}^{u}_{cross} \in \mathbb{R} ^ {(C+1)\times W\times H\times L}$. The loss functions for labeled and unlabeled images are defined as
\begin{align}
\label{eq:loss_cross}
    &\mathcal{L}_{cross}^l(\mathcal{B};\mathbf{\Theta}^s) = \ell_{dice}({\textbf{P}}^{l}_{cross},\textbf{Y}^l), \\
    &\mathcal{L}_{cross-in}^u(\mathcal{B};\mathbf{\Theta}^s) = \ell_{dice}({\textbf{P}}^{u}_{cross},\widehat{\textbf{Y}}^u).
\end{align}
where $\ell_{dice}$ denotes multi-class Dice loss. $\textbf{Y}^l\in\{0,1,...,C\}^{W\times H\times L}$ denotes per-voxel manual annotations for $\textbf{X}^l$. $\mathcal{L}_{cross-in}^u$ indicates the refined pseudo-labels for $\textbf{X}^u$ by blending cube-wise representations and more details can be seen in Sec.~\ref{sec:blending}.

\noindent\textbf{Within-image Partition and Recovery}
\label{within-image branch}
Besides magic-cubes partition and recovery across images, we also design a within-image partition and recovery branch for single images, which can better consider local features and learn local attributes for identifying targets, especially targets of small sizes. 
For $\textbf{X}^l$, the partitioned $j$-th cube $\textbf{x}_{j}^l$ is fed into $\mathcal{F}(\cdot;\mathbf{\Theta}^s)$ and the softmax layer, which can acquire a cube-wise prediction map ${\textbf{p}}_{j}^l\in\mathbb{R}^{C\times W/N\times H/N\times L/N}$. Finally, we recover the magic-cube via mixing $N^3$ cube-wise prediction maps back to their original positions as illustrated in Fig. \ref{fig:magic-cube partition and recovery}, and the recovered probability map is denoted as ${\textbf{P}}^{l}_{in} \in\mathbb{R}^{C\times W\times H\times L}$. Similarly, for $\textbf{X}^u$, the probability map is denoted as ${\textbf{P}}^{u}_{in} \in\mathbb{R}^{C\times W\times H\times L}$. This operation is denoted as $\mathcal{A}^{in}_{rec}(\cdot)$. The whole process for obtaining ${\textbf{P}}^{l}_{in}$ and ${\textbf{P}}^{u}_{in}$ are:
\begin{align}
{\textbf{P}}^{l}_{in},{\textbf{P}}^{u}_{in}=\mathcal{A}_{rec}^{in}(\sigma(\mathcal{F}(\mathcal{A}_{part}(\textbf{X}^\mathcal{B});\mathbf{\Theta}^s))).
\end{align}
Since ${\textbf{P}}^{l}_{in}$ and ${\textbf{P}}^{u}_{in}$ are recovered via predictions of cubes, ${\textbf{P}}^{l}_{in}$ and ${\textbf{P}}^{u}_{in}$ are denoted as cube-wise representations in the rest of the paper.
The loss function for the labeled image is:
\begin{equation}
\label{eq:loss_in_l}
    \mathcal{L}_{in}^l(\mathcal{B};\mathbf{\Theta}^s)=\ell_{dice}(\textbf{P}_{in}^l, \textbf{Y}^l).
\end{equation}
For the unlabeled image, we propose to leverage the cube-wise representations by refining the pseudo-label instead of directly computing the loss function as in Eq.~\ref{eq:loss_in_l}, which is illustrated in Sec.~\ref{sec:blending}. We conduct ablation studies on various design choices of utilizing the cube-wise representations in the experiment part.

\subsection{Cube-wise Pseudo-label Blending} 
\label{sec:blending}
As mentioned in the previous section, medical imaging experts reveal that local attributes \emph{e.g.}, texture, luster and boundary smoothness are crucial elements for identifying targets such as tiny organs in medical images \cite{zhao2022cross}. Inspired by this, we propose a within-image partition and recovery module to learn cube-wise local representations. 
{Since the teacher network takes the original volume as
input, and pays more attention to learn large organs in the head class, voxels that actually belong to the \emph{tail} class can be incorrectly predicted as \emph{head} class voxels, due to the lack of local attributes learning. }To effectively increase the chance of the voxel predicted to the tail class, we design a cube-wise pseudo-label blending module, which blends the original pseudo-label with cube-wise features.

\begin{figure}[t]
\begin{center}
    \includegraphics[width=1\linewidth]{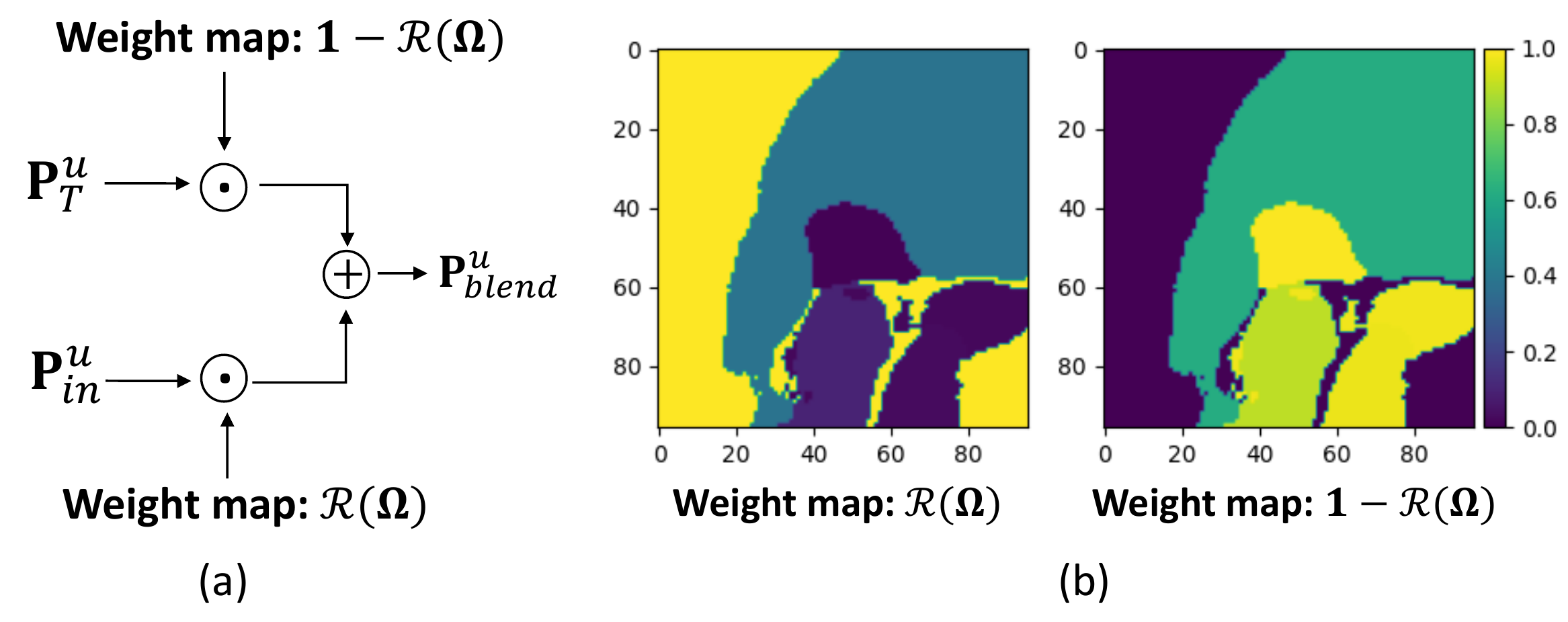}
\end{center}
\vspace{-1.3em}
\caption{
The cube-wise pseudo-label blending module. (a) Pipeline. (b) Illustration on $\mathcal{R}(\mathbf{\Omega}_\textbf{m})$. If the pseudo-label $\widetilde{\textbf{Y}}_\textbf{m}^u$ on $\textbf{m}$ is incorrectly assigned as a big organ, a big $\mathbf{\Omega}_\textbf{m}$ is assigned to ensure possible correction from cube-wise representations.} 
\label{fig:bld}
\vspace{-1.3em}
\end{figure}

In detail, the image-level prediction map of $\textbf{X}^u$ acquired from the teacher network is defined as $\textbf{P}_{T}^u=\mathcal{F}(\textbf{X}^u;\mathbf{\Theta}^t)$, where $\mathbf{\Theta}^t$ is the parameters of teacher network. The reconstructed cube-level representations of $\textbf{X}^u$ acquired from the student network is $\textbf{P}^u_{in}$. As mentioned above, cube-wise features pay more attention to learn local attributes which is important for tiny organs. Therefore, we propose a distribution-aware blending strategy, which is shown in Fig.~\ref{fig:bld}(a), and formulated as:
\begin{align}
\label{eq:score}
    \textbf{P}_{blend}^u=(\textbf{1}-\mathcal{R}(\mathbf{\Omega}))\odot \textbf{P}^u_T+\mathcal{R}(\mathbf{\Omega})\odot \textbf{P}_{in}^u,
\end{align}
where $\odot$ indicates the element-wise multiplication, $\textbf{1}\in\{1\} ^ {(C+1)\times W\times H\times L}$, and $\mathbf{\Omega}\in\mathbb{R} ^ {W\times H\times L}$ is a distribution-aware weight map. 

To obtain the weight map, we firstly learn the class-wise distribution $\textbf{v}$ during training. Suppose $\textbf{v}\in\mathbb{R}^C$ is a vector, $\textbf{v} = \{\textbf{v}_0,...,\textbf{v}_{C-1}\}$, whose element $\textbf{v}_c$ indicates the number of voxels belonging to $(c+1)$th organ, which is accumulated by counting the voxels \emph{w.r.t.} pseudo-labels over a few previous iterations. Let $\widetilde{\textbf{Y}}^u_\textbf{m}$ denote the pseudo-label for the voxel on location $\textbf{m}$. The value $\mathbf{\Omega}_{\textbf{m}}\in\mathbb{R}^1$at each spatial location $\textbf{m}$ on the weight map is derived as:
\begin{align}
\label{eq:class-distribution}
    \mathbf{\Omega}_{\textbf{m}}=\sum_{c=0}^{C-1}\frac{\mathbf{I}(\widetilde{\textbf{Y}}^u_\textbf{m}=c)\textbf{v}_c}{\max \textbf{v}}
\end{align}
where $\mathbf{I}(\cdot)$ is an indicator function. Note that to match with the data dimension in Eq.~\ref{eq:score}, we replicate $\mathbf{\Omega}$ $C+1$ times to acquire $\mathcal{R}(\mathbf{\Omega})\in\mathbb{R} ^ {(C+1)\times W\times H\times L}$. 

Overall, since the teacher network takes the original volume as input, and pays more attention to learn large organs in the head class, the teacher pseudo-label may be biased. 
To unbias the pseudo-label, on the one hand, we keep the pseudo-label of small organs. On the other hand, we remedy the possible incorrect pseudo-labels of big organs via changing them to small-organ classes via Eq.~\ref{eq:score} if $\mathbf{\Omega}_\textbf{m}$ is large. This can effectively increase the chance of the voxels predicted to small-organs.

The final refined pseudo-label $\widehat{\textbf{Y}}^u$ is obtained via $\widehat{\textbf{Y}}^u_\textbf{m}=\arg\max_{c\in \mathcal{C}}\textbf{P}^u_{blend,c,\textbf{m}}$, where $\mathcal{C}$ indicates the finite set of class labels, and $\textbf{m}$ means the voxel location.

\subsection{Magic-cube Location Reasoning}
To fully leverage the prior anatomical knowledge of multi-organs, we further propose a magic-cube location reasoning method, which is in line with within-image partition. For one image $\textbf{X}^l$, the partitioned $j$-th cube's $\textbf{x}_{j}^l$ is fed into the encoder of student network $\mathcal{F}_{enc}(\cdot; \mathbf{\Theta}_{enc}^s)$ to generate $\textbf{f}_{j}^l \in \mathbb{R}^{D\times W/N\times H/N\times L/N}$. Then, the feature $\textbf{f}_{j}^l$ is flattened into a one-dimensional vector, passing through a classification head $\mathcal{F}_{cls}(\cdot; \mathbf{\Theta}_{cls})$ to project the flattened feature vector into the same size as the number of cubes in one image, \emph{i.e.}, $N^3$. The classification head is composed of two fully connected layers \cite{Wei_2019_CVPR,fu2020domain}. Finally, we concatenate the output $\hat{\textbf{y}}_j^l \in \mathbb{R}^{N^3}$ from the classification head for all $N^3 $cubes in one image, which generates a vector in size $(N^3)^2$. This vector is further reshaped into a matrix of size $N^3 \times N^3$, whose rows indicate the probabilities of corresponding cubes belonging to the $N^3$ locations. The same applies to $\textbf{X}^u$. For both $\textbf{X}^l$ and $\textbf{X}^u$ in $\mathcal{B}$, the cross-entropy loss is adopted to learn relative locations of cubes\vspace{-0.7em}:

\begin{align}
\label{eq:cls}
    \mathcal{L}_{cls}(\mathcal{B};&\mathbf{\Theta}_{enc}^s,\mathbf{\Theta}_{cls})=\frac{1}{|\mathcal{B}|}\sum_{\textbf{X}\in\mathcal{B}} \ell_{ce}(\sigma(\mathcal{F}_{cls}\\\nonumber
    &(\mathcal{F}_{enc}^s(\mathcal{A}_{part}(\textbf{X});\mathbf{\Theta}_{enc}^s); \mathbf{\Theta}_{cls})),\mathbbm{y})
\end{align}
where $\ell_{ce}$ denotes the cross-entropy loss, $\mathbbm{y}$ represents cube's relative locations in $\textbf{X}$.

\subsection{Loss Function}
Overall, we develop a total loss function based on a mini-batch %$\mathcal{L}(\mathcal{B};\mathbf{\Theta}_{enc}^s, \mathbf{\Theta}_{seg}^s,\mathbf{\Theta}_{cls})$ 
for our {MagicNet} as follows:
\begin{align}
\label{eq:totalloss}
    \mathcal{L}(\mathcal{B};&\mathbf{\Theta}_{enc}^s, \mathbf{\Theta}_{seg}^s,\mathbf{\Theta}_{cls}) = \mathcal{L}_{sup}(\mathcal{B};\mathbf{\Theta}_{enc}^s,\mathbf{\Theta}^s,\mathbf{\Theta}_{cls}) \\\nonumber
    &+\mathcal{L}_{unsup}(\mathcal{B};\mathbf{\Theta}_{enc}^s,\mathbf{\Theta}^s,\mathbf{\Theta}_{cls}),
\end{align}
where $\mathcal{L}_{sup}(\cdot)$ and $\mathcal{L}_{unsup}(\cdot)$  denote the loss for labeled and unlabeled images in $\mathcal{B}$, respectively:
\begin{align}
\label{eq:sup}
    &\mathcal{L}_{sup}(\mathcal{B};\mathbf{\Theta}_{enc}^s,\mathbf{\Theta}^s,\mathbf{\Theta}_{cls})=\mathcal{L}_{cross}^{l}(\mathcal{B};\mathbf{\Theta}^s)\\\nonumber
    &+\mathcal{L}_{in}^l(\mathcal{B};\mathbf{\Theta}^s)+\beta\mathcal{L}_{cls}^l(\mathcal{B};\mathbf{\Theta}_{enc}^s,\mathbf{\Theta}_{cls}),
\end{align}
where $\beta$ is the weight factor to balance the location reasoning term with others, 
and
\begin{align}
\label{eq:unsup}
    \mathcal{L}_{unsup}(\mathcal{B};\mathbf{\Theta}_{enc}^s,\mathbf{\Theta}^s,&\mathbf{\Theta}_{cls})=\alpha\mathcal{L}_{cross-in}^{u}(\mathcal{B};\mathbf{\Theta}^s)\\\nonumber
    &+\beta\mathcal{L}_{cls}^{u}(\mathcal{B};\mathbf{\Theta}_{enc}^s,\mathbf{\Theta}_{cls}),
\end{align}
where $\alpha$ is another weight factor to balance the two terms.
For $\mathcal{L}_{cls}^l$ and $\mathcal{L}_{cls}^u$ in Eq.~\ref{eq:sup} and Eq.~\ref{eq:unsup}, the following equation holds:
\begin{equation}
    \mathcal{L}_{cls}=\mathcal{L}_{cls}^l+\mathcal{L}_{cls}^u.
\end{equation}

\subsection{Testing Phase}
Given a testing image $\textbf{X}_{test}$, the probability map is obtained by: 
    $\textbf{P}_{test}=\sigma(\mathcal{F}(\textbf{X}_{test}; \mathbf{\Theta}^{s*}))$,
where $\mathbf{\Theta}^{s*}$ denotes the well-trained student network parameters. The final label map can be determined by taking the \emph{argmax} in the first dimension of $\textbf{P}_{test}$.

\begin{table*}[!tb]
\renewcommand\arraystretch{0.93}
\small
\centering
\resizebox{0.84\linewidth}{!}{
\begin{tabular}{c|l|lllllllllllll|cc}
\toprule[0.15em]
\multirow{2}{*}{Labeled\%} & \multirow{2}{*}{Methods} & \multirow{2}{*}{Spl} & \multirow{2}{*}{R.kid} & \multirow{2}{*}{L.kid} & \multirow{2}{*}{Gall} & \multirow{2}{*}{Eso} & \multirow{2}{*}{Liv} & \multirow{2}{*}{Sto} & \multirow{2}{*}{Aor} & \multirow{2}{*}{IVC} & \multirow{2}{*}{Veins} & \multirow{2}{*}{Pan} & \multirow{2}{*}{RG} & \multirow{2}{*}{LG} & \multirow{2}{0.7cm}{\centering Avg. DSC} & \multirow{2}{0.7cm}{\centering Avg. NSD}\\
& & & & & & & & & & & & & & & &\\ 
\midrule[0.09em]
30\% & V-Net\cite{milletari2016v} & 86.37 & 75.38 & 71.20 & 41.81 & 48.46 & 92.77 & 44.78 & 88.20 & 72.60 & 49.27 & 22.30 & 37.79 & 21.71 & 57.90 & 56.90\\
& MT\cite{NIPS2017_meanteacher}& 79.89 & 77.56 & 78.08 & 38.31 & 58.99 & 92.26 & 48.73 & 88.61 & 79.36 & 52.73 & 28.42 & 54.16 & 21.30 & 61.42 & 60.85\\
& UA-MT\cite{Yu2019uncertainty} & 87.75 & 73.87 & 72.97 & 42.08 & 50.79 & 93.22 & 42.07 & 88.30 & 71.80 & 55.27 & 36.61 & 43.22 & 37.67 & 61.20 & 60.99\\
& ICT\cite{ijcai2019-504} & 88.78 & 78.00 & 83.93 & 50.09 & 53.35 & 88.32 & 64.19 & 89.02 & 79.60 & 63.51 & 58.90 & 43.29 & 50.70 & 68.59 & 67.77\\
& CPS\cite{chen2021semi} & 82.98 & 76.91 & 79.21 & 29.91 & 52.53 & 92.59 & 41.37 & 86.59 & 74.73 & 49.53 & 18.96 & 45.13 & 24.06 & 58.04 & 56.64\\
& SS-Net\cite{wu2022exploring} & 85.53 & 73.32 & 80.70 & 30.85 & 52.44 & 91.67 & 17.67 & 86.28 & 72.39 & 47.40 & 22.95 & 37.38 & 39.01 & 56.74 & 54.38\\
& SLC-Net\cite{liu2022slcnet} & 90.04 & 84.38 & 85.84 & 53.55 & 55.87 & 92.86 & 57.80 & 89.83 & 80.87 & 53.22 & 38.78 & 48.10 & 37.03 & 66.78 & 68.81\\
& \textbf{MagicNet (ours)} & \textbf{91.42} & \textbf{84.64} & \textbf{86.19} & \textbf{62.86} & \textbf{62.49} & \textbf{93.89} & \textbf{72.87} & \textbf{90.70} & \textbf{83.52} & \textbf{70.07} & \textbf{64.94} & \textbf{60.88} & \textbf{57.48} & \textbf{75.53} & \textbf{76.31}\\
\cmidrule{1-17}
40\% & V-Net\cite{milletari2016v} & 84.98 & 82.72 & 82.07 & 36.64 & 63.48 & 93.54 & 57.49 & 89.74 & 78.63 & 60.42 & 49.39 & 55.60 & 38.49 & 67.17 & 67.84\\
& MT\cite{NIPS2017_meanteacher} & 85.70 & 78.93 & 79.08 & 42.80 & 61.09 & 93.45 & 57.57 & 89.70 & 80.30 & 63.95 & 41.14 & 50.46 & 29.69 & 65.68 & 65.98\\
& UA-MT\cite{Yu2019uncertainty} & 88.74 & 75.88 & 78.91 & 54.25 & 58.55 & 93.46 & 58.90 & 89.23 & 76.15 & 62.30 & 47.91 & 51.53 & 44.92 & 67.75 & 68.87\\
& ICT\cite{ijcai2019-504} & 90.31 & 84.41 & 86.96 & 49.22 & 65.65 & 94.29 & 65.95 & 90.23 & 81.44 & 69.56 & 66.61 & 57.35 & 56.01 & 73.69 & 74.98\\
& CPS\cite{chen2021semi} & 87.56 & 72.99 & 77.59 & 53.31 & 54.08 & 92.41 & 54.58 & 87.75 & 74.32 & 58.68 & 48.02 & 50.39 & 43.86 & 65.81 & 65.34\\
& SS-Net\cite{wu2022exploring} & 84.74 & 76.37 & 74.19 & 43.42 & 57.05 & 92.90 & 14.37 & 83.14 & 69.77 & 52.45 & 27.08 & 54.29 & 27.66 & 58.26 & 57.75\\
& SLC-Net\cite{liu2022slcnet} & 90.05 & 84.00 & 86.43 & 56.16 & 58.91 & \textbf{94.68} & 70.72 & 89.93 & 79.45 & 60.59 & 54.22 & 51.03 & 39.08 & 70.40 & 72.87\\

& \textbf{MagicNet (ours)} & \textbf{91.61} & \textbf{85.02} & \textbf{88.13} & \textbf{58.16} & \textbf{66.72} & 94.07 & \textbf{74.46} & \textbf{90.77} & \textbf{84.31} & \textbf{71.56} & \textbf{68.90} & \textbf{63.48} & \textbf{60.47} & \textbf{76.74} & \textbf{78.68}\\
\cmidrule{1-17}
100\% & V-Net\cite{milletari2016v} & 84.00 & 84.82 & 86.38 & 67.42 & 65.02 & 94.83 & 73.75 & 90.27 & 84.19 & 69.85 & 63.54 & 62.60 & 65.02 & 76.28 & 77.45\\
\bottomrule[0.15em]
\end{tabular}
}
\caption{Comparison results (DSCs of each organ, avg. DSC, and avg. NSD) between our method and existing semi-supervised medical image segmentation methods on the \emph{\textbf{BTCV dataset}}. 
\textbf{V-Net} means training with only labeled samples on the V-Net backbone. \textbf{DSC}: Dice Similarity Coefficient. \textbf{NSD}: Normalized Surface Dice. The standard deviations are not reported due to space limit. Note: Spl: spleen, R.Kid: right kidney, L.Kid: left kidney, Gall: gallbladder, Eso: esophagus, Liv: liver, Sto: stomach, Aor: aorta, IVC: inferior vena cava, Veins: portal and splenic veins, Pan: pancreas, LG/RG: left/right adrenal glands.}
\label{comparison_btcv}
\end{table*}

\begin{table*}[!tb]
\renewcommand\arraystretch{0.92}
\small
\centering
\resizebox{0.84\linewidth}{!}{
\begin{tabular}{c|l|llllllll|cc}
\toprule[0.15em]
\multirow{2}{*}{Labeled\%} & \multirow{2}{*}{Methods} & \multirow{2}{*}{Spleen}& \multirow{2}{*}{\centering L.kidney} & \multirow{2}{*}{Gallbladder} & \multirow{2}{*}{Esophagus} & \multirow{2}{*}{Liver} & \multirow{2}{*}{Stomach} & \multirow{2}{*}{Pancreas} & \multirow{2}{*}{Duodenum} & \multirow{2}{0.8cm}{\centering Avg. DSC} & \multirow{2}{0.8cm}{\centering Avg. NSD} \\ 
& & & & & & & & & & &\\ 
\midrule[0.09em]
10\% & V-Net\cite{milletari2016v} & 88.89 \scriptsize{{(16.31})} & 84.90 \scriptsize{{(20.11})} & 55.73 \scriptsize{{(34.27})} &  
58.27 \scriptsize{{(20.05})} &  
93.15 \scriptsize{{(7.11})} &  
57.15 \scriptsize{{(30.62})} &
53.33 \scriptsize{{(22.44})} &
35.28 \scriptsize{{(19.83})} & 
65.84& 49.96\\

&MT\cite{NIPS2017_meanteacher} & 89.22 \scriptsize{{(13.56})} & 88.04 \scriptsize{{(17.61})} & 61.40 \scriptsize{{(31.91})} &  
58.89 \scriptsize{{(20.62})} &  
93.60 \scriptsize{{(5.04})} &  
74.00 \scriptsize{{(20.37})} &
64.70 \scriptsize{{(19.31})} &
44.53 \scriptsize{{(17.61})} & 
71.80& 53.54\\

&UA-MT\cite{Yu2019uncertainty} & 89.94 \scriptsize{{(13.36})} & 89.26 \scriptsize{{(15.67})} & 59.19 \scriptsize{{(32.35})} &  
59.43 \scriptsize{{(19.45})} &  
93.77 \scriptsize{{(5.05})} &  
75.43 \scriptsize{{(19.61})} &
65.86 \scriptsize{{(17.48})} &
44.57 \scriptsize{{(17.40})} & 
72.18& 54.04\\

&ICT\cite{ijcai2019-504} & 90.27 \scriptsize{{(13.70})} & 89.89 \scriptsize{{(15.81})} & 61.47 \scriptsize{{(30.31})} & 58.70 \scriptsize{{(20.17})} &  
93.47 \scriptsize{{(5.39})} &  
72.45 \scriptsize{{(20.95})} &
65.81 \scriptsize{{(18.28})} &
45.00 \scriptsize{{(19.22})} & 
72.13& 54.07\\

&CPS\cite{chen2021semi} & 91.73 \scriptsize{{(12.11})} & 90.59 \scriptsize{{(14.85})} & 60.54 \scriptsize{{(32.29})} &  
55.82 \scriptsize{{(21.03})} &  
94.19 \scriptsize{{(4.93})} &  
78.80 \scriptsize{{(15.28})} &
62.93 \scriptsize{{(20.03})} &
32.58 \scriptsize{{(18.68})} & 
70.90& 52.80\\

&SS-Net\cite{wu2022exploring} & 83.52 \scriptsize{{(19.56})} & 78.28 \scriptsize{{(25.05})} & 53.08 \scriptsize{{(33.88})} &  
56.24 \scriptsize{{(20.54})} &  
91.84 \scriptsize{{(6.07})} &  
47.65 \scriptsize{{(29.70})} &
43.23 \scriptsize{{(24.11})} &
25.70 \scriptsize{{(18.71})} & 
59.94& 42.68\\

&SLC-Net\cite{liu2022slcnet} & 86.89 \scriptsize{(17.54)} & 87.54 \scriptsize{(18.00)} & 55.28 \scriptsize{(34.75)} & 58.51 \scriptsize{(20.54)} & 93.70 \scriptsize{(5.33)} & 69.74 \scriptsize{(22.89)} & 46.55 \scriptsize{(26.21)} & 28.86 \scriptsize{(18.87)} & 65.89 & 13.28\\

&\textbf{MagicNet (ours)} & \textbf{92.33} \scriptsize{{(11.12})} & \textbf{91.19} \scriptsize{{(14.18})} & \textbf{66.35} \scriptsize{{(29.29})} &  
\textbf{70.95} \scriptsize{{(12.30})} &  
\textbf{94.31} \scriptsize{{(4.64})} &  
\textbf{81.56} \scriptsize{{(14.91})} &
\textbf{76.31} \scriptsize{{(10.56})} &
\textbf{61.50} \scriptsize{{(12.85})} & 
\textbf{79.31}& \textbf{62.31}\\

\midrule[0.05em]
20\% & V-Net\cite{milletari2016v} & 90.53 \scriptsize{{(15.33})} & 88.28 \scriptsize{{(19.76})} & 65.16 \scriptsize{{(30.03})} &  
66.10 \scriptsize{{(16.98})} &  
94.49 \scriptsize{{(4.18})} &  
79.56 \scriptsize{{(16.56})} &
69.04 \scriptsize{{(17.43})} &
51.98 \scriptsize{{(18.53})} & 
75.64& 59.66\\

&MT\cite{NIPS2017_meanteacher} & 91.76 \scriptsize{{(12.77})} & 91.40 \scriptsize{{(14.49})} & 63.83 \scriptsize{{(31.74})} &  
64.15 \scriptsize{{(18.05})} &  
94.44 \scriptsize{{(4.86})} &  
81.19 \scriptsize{{(15.79})} &
72.24 \scriptsize{{(15.79})} &
57.01 \scriptsize{{(16.70})} & 
77.00& 60.32\\

&UA-MT\cite{Yu2019uncertainty} & 92.39 \scriptsize{{(11.31})} & 91.44 \scriptsize{{(14.61})} & 63.63 \scriptsize{{(31.46})} &  
64.33 \scriptsize{{(17.92})} &  
94.50 \scriptsize{{(4.77})} &  
83.21 \scriptsize{{(14.03})} &
71.93 \scriptsize{{(15.51})} &
57.21 \scriptsize{{(17.35})} & 
77.33& 60.64\\

&ICT\cite{ijcai2019-504} & 92.96 \scriptsize{{(9.60})} & 91.67 \scriptsize{{(14.52})} & 65.90 \scriptsize{{(30.49})} &  
64.35 \scriptsize{{(16.79})} &  
94.46 \scriptsize{{(4.66})} &  
82.60 \scriptsize{{(15.46})} &
74.18 \scriptsize{{(12.99})} &
57.67 \scriptsize{{(16.16})} & 
77.97& 61.24\\

&CPS\cite{chen2021semi} & 92.66 \scriptsize{{(11.46})} & 91.87 \scriptsize{{(14.33})} & 64.75 \scriptsize{{(31.21})} &  
57.27 \scriptsize{{(19.56})} &  
\textbf{94.97} \scriptsize{{(4.51})} &  
85.65 \scriptsize{{(9.50})} &
74.15 \scriptsize{{(12.49})} &
55.00 \scriptsize{{(16.58})} & 
77.04& 60.01\\

&SS-Net\cite{wu2022exploring} & 91.17 \scriptsize{{(14.77})} & 87.77 \scriptsize{{(19.37})} & 63.49 \scriptsize{{(30.83})} &  
65.02 \scriptsize{{(16.65})} &  
93.76 \scriptsize{{(4.82})} &  
73.04 \scriptsize{{(22.16})} &
70.37 \scriptsize{{(17.70})} &
52.81 \scriptsize{{(20.32})} & 
74.68& 57.57\\

&SLC-Net\cite{liu2022slcnet} & 92.60 \scriptsize{{(11.24})} & 91.38 \scriptsize{{(14.57})} & 62.46 \scriptsize{{(31.37})} &  
62.84 \scriptsize{{(19.31})} &  
94.54 \scriptsize{{(4.72})} &  
80.30 \scriptsize{{(14.73})} &
69.79 \scriptsize{{(16.99})} &
53.48 \scriptsize{{(18.40})} & 
75.92& 59.93\\

&\textbf{MagicNet (ours)} & \textbf{93.52} \scriptsize{{(10.22})} & \textbf{92.01} \scriptsize{{(14.30})} & \textbf{71.04} \scriptsize{{(26.92})} &  
\textbf{70.95} \scriptsize{{(12.58})} &  
94.89 \scriptsize{{(4.42})} &  
\textbf{85.77} \scriptsize{{(10.27})} &
\textbf{78.81} \scriptsize{{(8.07})} &
\textbf{66.21} \scriptsize{{(12.33})} & 
\textbf{81.65}& \textbf{65.87}\\

\midrule[0.05em]

100\% & V-Net\cite{milletari2016v} & 94.35 \scriptsize{{(5.81})} & 92.30 \scriptsize{{(14.22})} & 73.61 \scriptsize{{(28.80})} &  
73.48 \scriptsize{{(12.43})} &  
95.17 \scriptsize{{(3.10})} &  
89.09 \scriptsize{{(7.74})} &
80.86 \scriptsize{{(7.38})} &
70.03 \scriptsize{{(11.60})} & 
83.61 & 68.85\\

\bottomrule[0.15em]
\end{tabular}
}
\caption{Comparison results (DSCs of each organ, avg. DSC, and avg. NSD) between our method and existing semi-supervised medical image segmentation methods on \emph{\textbf{MACT dataset}} under 4-fold cross-validation. The value in ($\cdot$) is standard deviation. }
\label{comparison_mact}
\vspace{-0.7em}
\end{table*}

\section{Experiments}

\subsection{Datasets and Pre-processing}

\noindent\textbf{BTCV Multi-organ Segmentation Dataset} 
BTCV multi-organ segmentation dataset is from \emph{MICCAI Multi-Atlas Labeling Beyond Cranial Vault-Workshop Challenge}\cite{landman2015miccai} which contains 30 subjects with 3779 axial abdominal CT slices with 13 organs annotation. In pre-processing, we follow \cite{tang2022self} to re-sample all the CT scans to the voxel spacing $[1.5\times 1.5\times 2.0] mm^3$ and normalize them to have zero mean and unit variance. Strictly following \cite{fu2020domain,chen2021transunet,cao2021swinunet}, 18 cases are divided for training and the other 12 cases are for testing.

\noindent\textbf{MACT Dataset} 
Multi-organ abdominal CT reference standard segmentations (MACT) is a public multi-organ segmentation dataset, containing 90 CT volumes with 8 organs annotation. It is manually re-annotated by \cite{densevnet2018TMI} based on 43 cases of NIH pancreas \cite{roth2015deeporgan} and 47 cases of Synapse \cite{landman2015miccai}. We follow the setting of UMCT \cite{Xia2020uncertainty}. 1) In preprocessing, we set the soft tissue CT window range of $[-125, 275]$ HU and all the CT scans are re-sampled to an isotropic resolution of $1.0 mm^3$. Then, image intensities are normalized to have zero mean and unit variance. 2) We randomly split the dataset into four folds, and perform 4-fold cross-validation. Labeled training set is then randomly selected, which are applied to all experiments for fair comparison\vspace{-0.1em}.

\begin{table*}[!tb]
\renewcommand\arraystretch{0.9}
\small
\centering
\resizebox{0.9\linewidth}{!}{
\begin{tabular}{l|cccc|c|cccccccc}
\toprule[0.15em]
\multirow{2}{*}{\centering Methods} & \multirow{2}{0.6cm}{\centering Cross} & \multirow{2}{0.6cm}{\centering In} & \multirow{2}{0.6cm}{\centering Loc} & \multirow{2}{0.6cm}{\centering Bld} & \multirow{2}{1.2cm}{\centering Avg. DSC} & \multirow{2}{1.2cm}{\centering spleen} & \multirow{2}{1.2cm}{\centering l.kidney} & \multirow{2}{1.2cm}{\centering gallbladder} & \multirow{2}{1.2cm}{\centering esophagus} & \multirow{2}{1.2cm}{\centering liver} & \multirow{2}{1.2cm}{\centering stomach} & \multirow{2}{1.2cm}{\centering pancreas} & \multirow{2}{1.2cm}{\centering duodenum}\\
& & & & & & & & & & & & &\\ 
\midrule[0.09em]
Baseline (MT\cite{NIPS2017_meanteacher}) & & & & & 71.88 & 86.84 & 88.92 & 54.50 & 53.40 & 93.01 & 70.86 & 71.77 & 55.76\\
Cross & \checkmark & & & & 78.13 & 92.11 & 90.53 & 63.42 & 70.05 & 93.90 & 79.60 & 75.58 & 59.87\\
Cross + In& \checkmark & \checkmark & & & 78.83 & 92.82 & 90.75 & 64.91 & 70.68 & 93.90 & 80.92 & 76.25 & 60.40\\
Cross + Loc& \checkmark & & \checkmark & & 78.56 & 92.49 & 91.07 & 65.95 & 69.72 & 94.12 & 80.40 & 75.82 & 58.94\\
Cross + In + Loc & \checkmark & \checkmark & \checkmark & & 78.87 & 92.55 & 91.18 & 65.72 & 70.77 & 94.42 & 80.05 & 75.81 & 60.43\\
Cross + In + Loc + Bld & \checkmark & \checkmark & \checkmark & \checkmark & 79.31 & 92.33 & 91.19 & 66.35 & 70.95 & 94.31 & 81.56 & 76.31 & 61.50\\
\bottomrule[0.15em]
\end{tabular}
}
\caption{Ablation study (DSC \%) for the effectiveness of each component of {MagicNet}. \textbf{Cross}: cross-image partition-and-recovery. \textbf{In}: Within-image partition-and-recovery. \textbf{Bld}: cube-wise pseudo-label blending. \textbf{Loc}: magic-cube location reasoning. 
}
\label{ablation_study_total}
\vspace{-0.1em}
\end{table*}

\vspace{-0.17em}
\subsection{Experimental Setup and Evaluation Metrics\vspace{-0.1em}}
In this work, we conduct all the experiments based on PyTorch with one NVIDIA 3090 GPU and use V-Net 
%and 3D U-Net 
as our backbone\cite{Yu2019uncertainty,wu2022exploring,liu2022slcnet}. For magic-cube location reasoning, we add a classification head composed of two fully-connected layers at the end of our encoder. Firstly, for BTCV dataset, the framework is trained by an SGD optimizer for 70K iterations, with an initial learning rate (lr) 0.01 with a warming-up strategy: $lr = base\_lr \times (1 - \frac{iterations}{max\_iterations})^{0.9}$. For MACT dataset, the framework is trained by an SGD optimizer for 30K iterations, with an initial learning rate (lr) 0.01 decayed by 0.1 every 12K iterations.
Following \cite{Xia2020uncertainty,Luo2021DTC,Yu2019uncertainty,wu2022exploring,liu2022slcnet}, for both two datasets, batch size is set as 4, including 2 labeled images and 2 unlabeled images, and the hyper-parameter $\alpha$ is set using a time-dependent Gaussian warming-up function and $\beta$ is empirically set as 0.1\cite{fu2020domain}. 
%Following \cite{Xia2020uncertainty, Luo2021DTC}, 
We randomly crop volume of size $96^3$, and then we partition the volume into $N^3$ small cubes, forming our magic-cubes, and the choice of $N$ is also discussed in Sec. \ref{sec:N}. In the testing phase, a sliding window strategy is adopted to obtain final results with a stride of $16^3$. The final evaluation metrics for our method are DSC (\%, Dice-S{\o}rensen Coefficient) and NSD (\%, Normalized Surface Dice) which are commonly used in the current multi-organ segmentation challenge \cite{ji2022amos,Ma-2021-AbdomenCT-1K}.

\begin{figure}[t]
\begin{center}
    \includegraphics[width=1\linewidth]{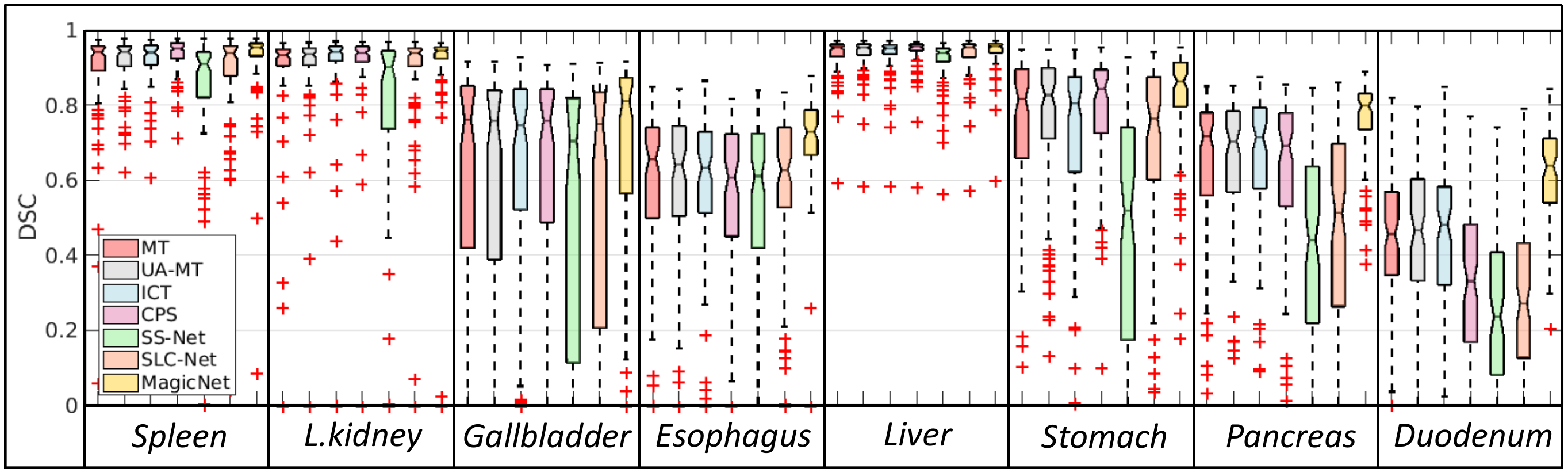}
\end{center}
\vspace{-0.7em}
\caption{DSC comparison in box plots on MACT dataset with 10\% labeled images (Best viewed electronically, zoom in).} 
\label{fig:boxplot}
\vspace{-1em}
\end{figure}

\subsection{Comparison with the State-of-the-art Methods}
\vspace{-0.3em}
We compare {MagicNet} with six state-of-the-art semi-supervised segmentation methods, including
mean-teacher (MT) \cite{NIPS2017_meanteacher}, uncertainty-aware mean-teacher (UA-MT) \cite{Yu2019uncertainty}, interpolated consistency training (ICT) \cite{ijcai2019-504}, cross pseudo-supervision (CPS) \cite{chen2021semi}, smoothness and class-separation (SS-Net) \cite{wu2022exploring}, shape-awareness and local constraints (SLC-Net) \cite{liu2022slcnet}. All experiments of other methods are re-trained by their official code when transferred to multi-organ task. V-Net is adopted as our backbone which is the same as the above methods for fair comparison.

In Table \ref{comparison_btcv}, we summarize the results of BTCV dataset. {Visualizations compared with state-of-the-arts are shown in the supplementary material.} Noticeable improvements compared with state-of-the-arts can be seen for organs such as {Stomach ($\uparrow$ 8.68), Gallbladder ($\uparrow$ 9.31), Veins ($\uparrow$ 6.56), Right adrenal glands ($\uparrow$ 6.72) and Left adrenal glands ($\uparrow$ 6.78).} It is interesting to observe that some semi-supervised methods even perform worse than the lower bound. This is because these methods are less able to learn common semantics from the labeled images to unlabeled ones,  generating less accurate pseudo-labels, which even hamper the usage of unlabeled images (see the supplementary material).

\begin{table*}[t]
\begin{minipage}[t]{0.7\columnwidth}
\renewcommand\arraystretch{0.9}
\footnotesize
\centering
\resizebox{1\linewidth}{!}{
\begin{tabular}{cc|cc}
\toprule[0.15em]
\ding{172} & \ding{173} & \multicolumn{1}{c}{DSC} & \multicolumn{1}{c}{NSD}\\
\midrule
scramble & U & 69.53 ~$\pm$~ 11.27 & 50.98 ~$\pm$~ 11.00\\
keep & U & 73.84  ~$\pm$~ 9.96 & 55.80 ~$\pm$~ 10.81\\
scramble & LU & 71.91 ~$\pm$~ 11.20 & 54.10 ~$\pm$~ 11.28\\
keep & LU & 78.13 ~$\pm$~ 8.12 & 60.79 ~$\pm$~ 9.90\\
\bottomrule[0.15em]
\end{tabular}
}
\caption{Ablation of design choices for cross-image partition and recovery (Question \ding{172} and \ding{173}, mean $\pm$ std of all cases). \textbf{Scramble}/\textbf{keep}: ignore/keep original positions when mixed. \textbf{U}: only for unlabeled data. \textbf{LU}: for both labeled and unlabeled data. The last row is ours.
}
\label{ablation_study_design_cross}
\end{minipage}
~~
\begin{minipage}[t]{0.67\columnwidth}
\renewcommand\arraystretch{1.1}

\centering
\resizebox{1\linewidth}{!}{
\begin{tabular}{c|cc}
\toprule[0.15em]
& \multicolumn{1}{c}{DSC} & \multicolumn{1}{c}{NSD}\\
\midrule
teacher sup. & 78.47 ~$\pm$~ 7.83 & 61.52 ~$\pm$~ 9.21\\
mutual sup. & 70.45 ~$\pm$~ 11.47 & 53.71 ~$\pm$~ 11.78\\
blending & 79.31 ~$\pm$~ 7.55 & 62.31 ~$\pm$~ 9.08\\
\bottomrule[0.15em]
\end{tabular}
}
\caption{Comparison of different pseudo-label supervision/blending strategies for unlabeled data. 
\textbf{Sup.}: supervision. \textbf{Blending}: our cube-wise pseudo-label blending. 
}
\label{ablation_study_design_blending}
\end{minipage}
~~
\begin{minipage}[t]{0.65\columnwidth}
\renewcommand\arraystretch{0.9}
\footnotesize
\centering
\resizebox{1\linewidth}{!}{
\begin{tabular}{c|cc}
\toprule[0.15em]
& \multicolumn{1}{c}{DSC} & \multicolumn{1}{c}{NSD}\\
\midrule
CutMix\cite{Yun_2019_ICCV} & 73.80 ~$\pm$~ 9.99 & 55.86 ~$\pm$~ 10.54\\
CutOut\cite{Terrance2017Cutout} & 72.47 ~$\pm$~ 10.27 & 54.62 ~$\pm$~ 10.67\\
MixUp\cite{zhang2018mixup} & 72.13 ~$\pm$~ 11.32 & 54.07 ~$\pm$~ 12.01\\
Ours (2) & 78.68 ~$\pm$~ 7.45 & 60.78 ~$\pm$~ 9.12\\
Ours (3) & 79.31 ~$\pm$~ 7.55 & 62.31 ~$\pm$~ 9.08\\
\bottomrule[0.15em]
\end{tabular}
}
\caption{Comparison of different data augmentation methods, where we try different CutMix and CutOut sizes, and choose the best results. For MagicNet, we compare different $N$ values in $(\cdot)$\vspace{-1em}.
}
\label{ablation_study_different_data_augmentations}
\end{minipage}
\end{table*}

We then evaluate MagicNet on MACT dataset. In Table \ref{comparison_mact}, we can observe: (1) a more significant performance improvement between {MagicNet} and other state-of-the-arts with 10\% labeled data than 20\%, which demonstrates the effectiveness of {MagicNet} when training with fewer labeled CT images; (2) MagicNet successfully mitigates the inferior performance in tail classes, \emph{e.g.}, 
DSC improvement with 10\% labeled images on {Gallbladder ($\uparrow$ 4.88), Esophagus ($\uparrow$ 11.52), Pancreas ($\uparrow$ 10.45) and Duodenum ($\uparrow$ 16.50).} Fig.~\ref{fig:boxplot} shows a comparison in DSC of six methods for different organs. {Note that UMCT \cite{Xia2020uncertainty} on MACT dataset achieves 75.33\% in DSC. It uses AH-Net \cite{liu20183d} as the backbone (3 are used), whose parameter number is much larger. Due to computation resources, we do not directly compare with it.}

\vspace{-0.4em}
\subsection{Ablation Study}
\vspace{-0.3em}
We conduct ablation studies on MACT dataset\cite{densevnet2018TMI} with 10\% labeled data to show the effectiveness 
of each module.

\noindent\textbf{The effectiveness of each component in MagicNet}
We conduct ablation studies to show the impact of each component in MagicNet. In Table \ref{ablation_study_total}, the first row indicates the mean-teacher baseline model \cite{NIPS2017_meanteacher}, which our method is designed on. Compared to the baseline, our magic-cube partition and recovery can yield good segmentation performance. \textbf{Cross} and \textbf{In} represent our cross- and within-image partition-and-recovery, which increase the performance from 71.88$\%$ to 78.13$\%$ and 78.87$\%$, respectively. This shows the powerful ability with our specially-designed data augmentation method. \textbf{Loc} represents magic-cube location reasoning for within-image branch. We can see from the table that based on our cross-image branch, adding relative locations of small-cubes can achieve 78.56$\%$ in DSC, adding only within-image partition-and-recovery branch achieves 78.83$\%$, and adding both two branches leads to 78.87$\%$.
Finally, our proposed cube-wise pseudo label blending (short for \textbf{Bld} in Table \ref{ablation_study_total}) provides significant improvement to 79.31$\%$, for which we will make a more complete analysis later in the current section.

\begin{figure}[t]
\begin{center}
    \includegraphics[width=0.86\linewidth]{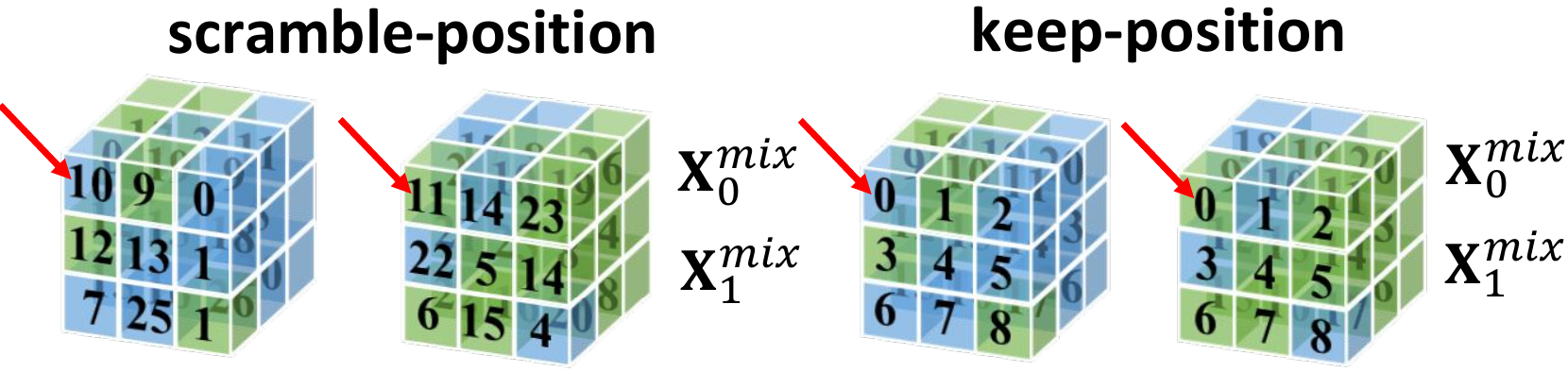}
\end{center}
\vspace{-0.8em}
\caption{Scramble position and keep position when mixing small-cubes across images (from Fig. \ref{fig:magic-cube partition and recovery}). Red arrow indicates the cube
location in mixed images.} 
\label{fig:position}
\vspace{-0.3em}
\end{figure}

\begin{figure}[t]
\begin{center}
    \includegraphics[width=1\linewidth]{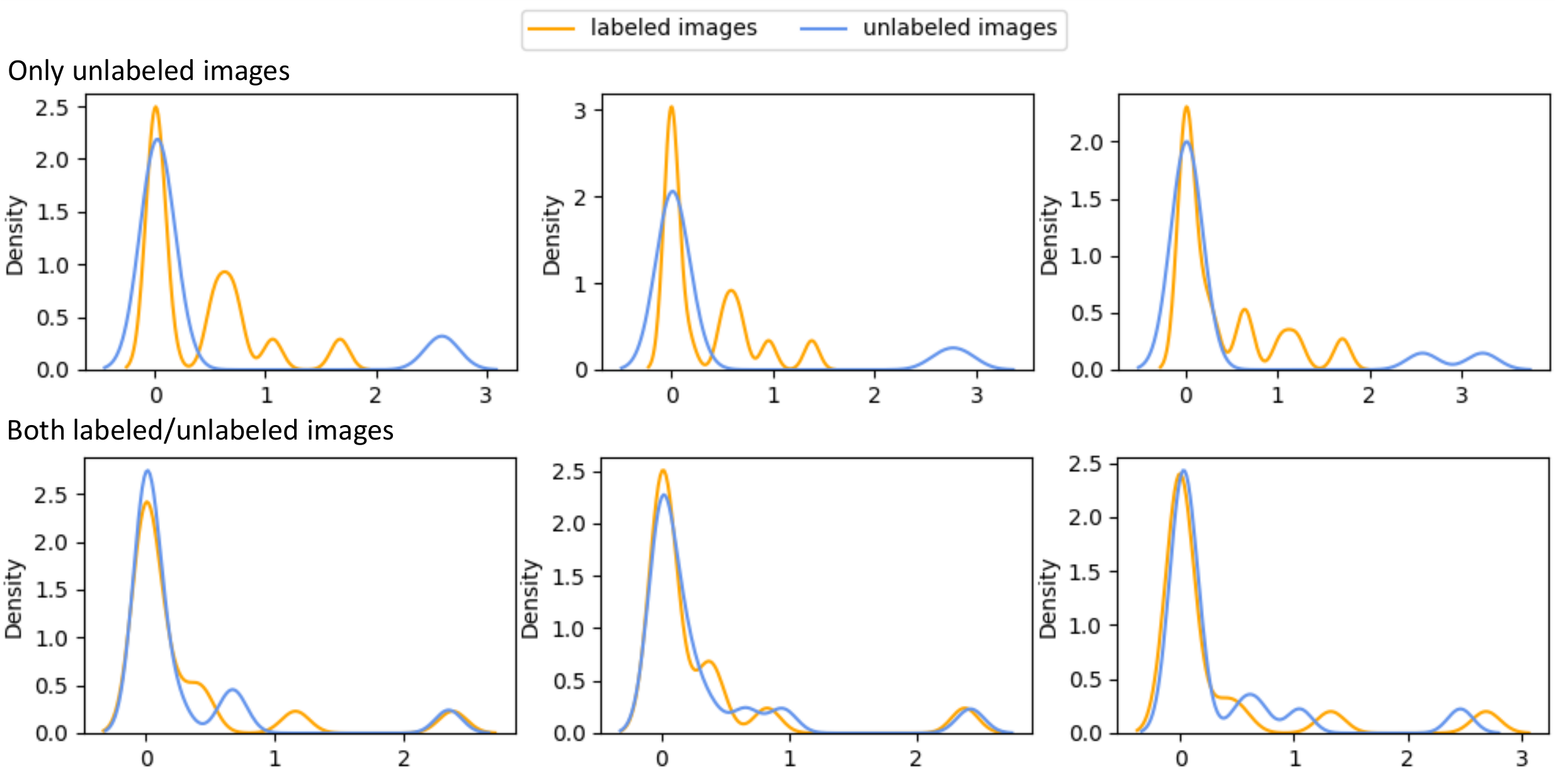}
\end{center}
\vspace{-1.3em}
\caption{
Kernel Density Estimation of labeled and unlabeled samples on the first three channels of the feature map of ``only unlabeled images" and ``both labeled and unlabeled images". Considerable distribution mismatch between labeled and unlabeled images can be observed (top), while our method can well improve the distribution mismatch (bottom). Best viewed zoom in.} 
\label{fig:kde}
\vspace{-1em}
\end{figure}

\noindent\textbf{Design choices of partition and recovery}
Here, we discuss the design for cross-image partition-and-recovery branch: \ding{172} Should we maintain or scramble the magic-cube relative locations when manipulating our cross-image branch, as shown in Fig.~\ref{fig:position}? The comparison results are shown in the last two rows in Table~\ref{ablation_study_design_cross}. \textbf{Scramble} and \textbf{Keep} represent the partitioned small-cubes are randomly mixed while ignoring their original locations or kept when mixing them across images. Results show that the relative locations between multiple organs are important for multi-organ segmentation.
\ding{173} Should our cross-image data augmentation be operated on only unlabeled images (see \textbf{U} in Table~\ref{ablation_study_design_cross}) or both labeled and unlabeled images (see \textbf{LU} in Table~\ref{ablation_study_design_cross})? The latter obtains a much better performance compared to the former. Besides, it largely reduces the distribution mismatch between labeled and unlabeled data, as shown in Fig.~\ref{fig:kde}.

\begin{figure}[t]
\begin{center}
    \includegraphics[width=0.62\linewidth]{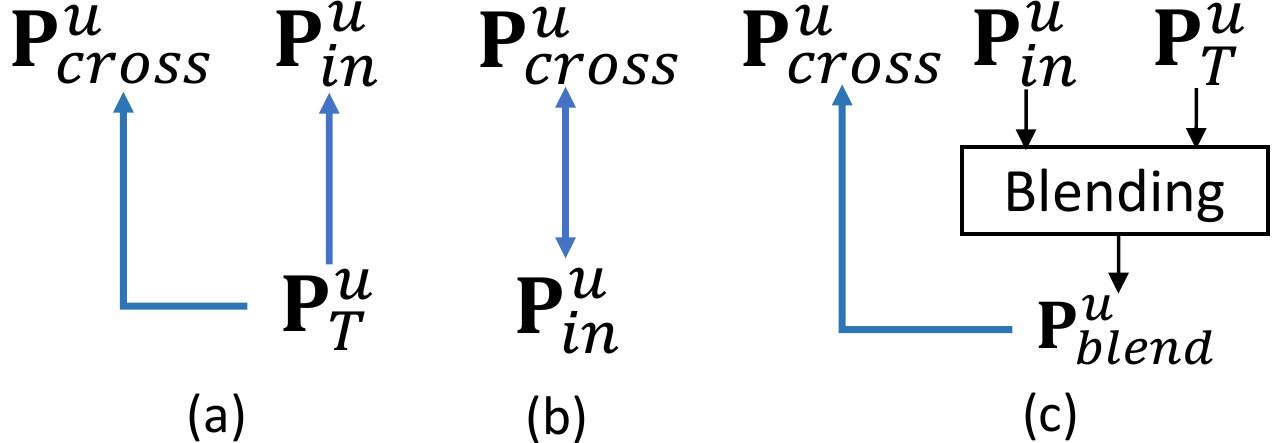}
\end{center}
\vspace{-0.8em}
\caption{Different pseudo-label supervision/blending strategies corresponding to Table \ref{ablation_study_design_blending}. \textbf{Blue line indicates supervision.} (a) teacher sup. (b) mutual sup. (c) our blending.} 
\label{fig:supervision}
\vspace{-0.5em}
\end{figure}

\noindent\textbf{Cube-wise pseudo-label blending}
As illustrated in Sec. \ref{sec:blending}, 
we blend the output of within-image branch and the output of teacher model to obtain the final pseudo-label for complementing local attributes. We compare our blending with two other methods, as shown in Table~\ref{ablation_study_design_blending}.  
Three supervision schemes, illustrated in Fig.~\ref{fig:supervision}, are compared for unlabeled images based on our framework. 
\textbf{Teacher sup.} means the outputs of cross- and within-image branch are both supervised by the pseudo-label from the teacher model. \textbf{Mutual sup.} means the outputs of cross- and within-image are mutually supervised. It can be seen that our blending strategy works favorably for unlabeled data.

\noindent\textbf{Comparison with interpolated-based methods}
As shown in Table~\ref{ablation_study_different_data_augmentations}, our augmentation method outperforms other methods such as CutMix, CutOut and MixUp. For fair comparison, we try several CutMix and CutOut sizes, and choose the best results. 

\noindent\textbf{Different number ($N$) of small-cubes}
\label{sec:N}
We study the impact of different numbers of small-cubes $N$, as shown in Table \ref{ablation_study_different_data_augmentations}. Slightly better performance is achieved when $N = 3$. When $N=4$, the size of the small cube does not match the condition for VNet. Thus, we only compare the results given $N=2$ and $N=3$.

\section{Limitations and Conclusion}

We have presented \emph{MagicNet} for semi-supervised multi-organ segmentation. MagicNet encourages unlabeled data
to learn organ semantics in relative locations from the labeled data, and enhances the learning ability for small organs. Two essential components are proposed: (1) partition-and-recovery $N^3$ small cubes cross- and within-labeled and unlabeled images, and (2) the quality of pseudo label refinement by blending the cube-wise representations. Experimental results on two public multi-organ segmentation datasets verify the superiority of our method.   

\noindent\textbf{Limitations:}  CT scans are all roughly aligned by the nature of the data. Magic-cube may not work well on domains \emph{w/o} such prior, \emph{e.g.}, microscopy images \emph{w/o} similar prior. 

\section{Acknowledgement}

This work was supported by the National Natural Science Foundation of China (Grant No. 62101191), Shanghai Natural Science Foundation (Grant No. 21ZR1420800), and the Science and Technology Commission of Shanghai Municipality (Grant No. 22DZ2229004).
%%%%%%%%% REFERENCES
\clearpage
{\small
\bibliographystyle{ieee_fullname}
\bibliography{egbib}
}

\end{document}